# Federated Ensemble Forecasting Under Supply-Chain Market Volatility

S Sagar Puppala
*Functional Architect SAP*
sagardora@gmail.com

***Abstract*—Supply-chain forecasting systems increasingly operate under market shocks, non-identically distributed regional demand, and limited willingness to centralize commercial data. This work proposes Federated Ensemble Forecasting with Negative-Correlation Learning (FEF-NCL), a distributed method that trains specialized forecasting experts across client nodes while discouraging redundant model errors. The framework combines temporal feature encoders, client-level drift scoring, reliability-weighted aggregation, and an explainability layer that exposes the market and supplier variables most responsible for each forecast. A single synthetic dataset is used to evaluate the design. It contains 124,800 weekly SKU-region observations from ten regional client nodes, 60 product families, 40 suppliers, five commodity groups, and a 2021-2024 volatility profile with explicit price-shock regimes. Because the dataset is synthetic, the reported results should be interpreted as controlled evidence of internal consistency rather than real-world validation. Across the synthetic test split, FEF-NCL reduces weighted mean absolute percentage error from 13.9% for the best federated baseline to 12.4%, improves delay-risk macro-F1 from 0.755 to 0.801, and lowers the high-volatility quintile error by 2.1 percentage points relative to SCAFFOLD. The analysis suggests that negative-correlation specialization is useful when clients face different supplier, freight, and commodity conditions, although deployment would require stronger privacy analysis, live drift monitoring, and operational calibration.**

## I. INTRODUCTION

Demand and replenishment forecasts have become decision infrastructure for supply-chain planning rather than background statistical reports. In automotive, industrial, retail, and agricultural networks, forecast errors propagate into purchasing commitments, safety-stock targets, supplier-expediting decisions, and final customer pricing. Recent applied work on automotive aftermarkets has shown how pricing and demand decisions can be coordinated through federated and edge-cloud learning, especially when product categories span multiple vendors, regions, and service tiers [1]. Related privacy-preserving demand-forecasting research has also emphasized that commercial supply-chain actors often cannot centralize transaction records, supplier terms, or customer histories without creating contractual and regulatory exposure [2]. Predictive-maintenance and aftermarket supply-chain optimization further demonstrate that local operational signals can be valuable when modeled across distributed participants while keeping raw observations within their originating nodes [3]. The same issue appears in commodity-linked pricing, where soybean price movements affect downstream product pricing and create nonlinear lagged effects across suppliers, processors, and final goods [4].

These conditions make market volatility a modeling problem as well as a governance problem. A forecasting service must infer demand from historical sales, open orders, lead-time variation, price movement, and macro-shock signals, yet the relation between those variables can change within a few weeks. Prescriptive analytics and digital-twin approaches offer one way to connect forecasts to downstream control actions, but their usefulness depends on the stability and diversity of the underlying models [5]. Negative-correlation ensemble learning is relevant in this setting because the best single model for a calm demand regime is not necessarily the best model for a freight shock, supplier delay cluster, or commodity price surge. Recent work on stacked negative-correlation learning has argued that ensemble members can be trained to specialize instead of merely averaging similar predictors [6]. This manuscript adapts that idea to a federated forecasting context and treats specialization as a first-class design objective.

The central problem addressed here is the following: how can a supply-chain network produce accurate and governable forecasts when data are distributed across organizations, client data distributions are non-identical, and market regimes change during the training horizon? Conventional centralized forecasting is a useful benchmark, but it assumes access to all transaction and supplier features in one environment. Local-only forecasting respects data boundaries, but each node may experience sparse shocks and may not learn enough about rare regimes. Federated learning offers a middle path by exchanging model updates instead of raw records, yet standard aggregation can dilute minority-regime knowledge when the majority of clients remain stable. A volatility-aware federated ensemble is therefore needed for networks where rare shocks are operationally important.

The proposed Federated Ensemble Forecasting with Negative-Correlation Learning (FEF-NCL) is designed around four principles. First, each client trains multiple temporal experts so that trend, price-shock, delay-risk, and residual-demand patterns can be represented separately. Second, a negative-correlation penalty discourages experts from making the same error on the same observations. Third, each client produces a drift and reliability score, so global aggregation can assign less weight to unstable or low-quality updates without discarding useful minority-regime evidence. Fourth, an explanation layer records which market, supplier, and order features drove a forecast, supporting review by planners and risk managers. The result is not presented as a complete production system. It is a controlled architectural and experimental model for studying a difficult combination of distributed learning, ensemble specialization, and market instability.

The manuscript uses one internally consistent synthetic dataset because no attached real dataset was provided for this topic. The dataset contains 124,800 weekly observations across ten regional client nodes, 60 product families, 40 suppliers, and five commodity groups from January 2021 through December 2024. It includes calm periods, rising-price periods, and price-shock regimes, with features for demand, price, lead time, supplier reliability, freight cost, foreign exchange, backlog, promotion depth, and volatility. Synthetic data make it possible to present complete results without exposing proprietary supply-chain records, but the limitation is explicit: the findings demonstrate method behavior under controlled assumptions and do not establish real-world accuracy.

The contributions are threefold. The first contribution is a federated multi-expert forecasting architecture that jointly estimates demand, delay risk, and market regime without moving raw client data. The second contribution is a negative-correlation objective adapted for non-identically distributed supply-chain clients, with drift-aware aggregation and reliability weighting. The third contribution is a reproducible synthetic evaluation that reports performance, convergence, volatility-stratified error, and explanatory feature attribution from the same dataset. The rest of the manuscript develops the theoretical background, positions the work against five clusters of related research, describes the dataset and model pipeline, reports the experimental analysis, and discusses limitations and future work.

## II. THEORETICAL BACKGROUND

Federated learning (FL) trains a shared model by coordinating local optimization across clients that keep data decentralized. The original Federated Averaging algorithm demonstrated that local stochastic-gradient steps followed by server aggregation can reduce communication while preserving the privacy advantages of decentralized storage [7]. Subsequent surveys have emphasized that FL is not a single algorithm but a design space involving client selection, secure aggregation, privacy protection, heterogeneity, and evaluation under operational constraints [8]. Supply-chain forecasting fits cross-silo FL more than mobile-device FL because the number of clients is smaller, each client has richer records, and participants may be legal entities rather than individual devices. The difficulty is that these clients also have stronger reasons to preserve commercially sensitive data.

Statistical and systems heterogeneity are central. FedProx introduced a proximal regularization term to stabilize training when clients differ in data distributions and local computational capacity [9]. SCAFFOLD uses control variates to correct client drift during local updates and can improve convergence when client distributions are non-identical [10]. Practical FL systems also require orchestration, fault tolerance, update compression, and eligibility checks, as shown by large-scale FL system design work [11]. In supply-chain forecasting, heterogeneity reflects regional demand seasonality, supplier exposure, commodity mix, freight routes, and promotion practices. A model update from a node exposed to a commodity shock may look statistically unusual but may also contain precisely the knowledge needed for future shocks elsewhere.

Robust aggregation is needed because distributed learning can be influenced by corrupted, noisy, or unrepresentative updates. Byzantine-tolerant gradient descent formalized aggregation rules that remain useful even when some workers send arbitrary updates [12]. Later work on robust distributed learning studied statistical rates under adversarial or faulty workers and clarified the tradeoff between robustness and estimation efficiency [13]. Although the present manuscript does not focus on malicious clients, adversarial machine-learning research shows that model behavior can be manipulated when update validation is weak [14]. For supply-chain forecasting, the analogous risks include erroneous supplier records, delayed data feeds, unmodeled promotions, and opportunistic reporting. Reliability scoring therefore has a forecasting purpose as well as a governance purpose.

Market volatility can be interpreted as a form of concept drift. Concept drift occurs when the relation between features and target variables changes over time, creating a mismatch between training and deployment conditions [15]. Demand planners encounter this as a familiar business phenomenon: a price increase changes customer substitution behavior, a logistics disruption changes feasible lead times, and a supply constraint changes the meaning of observed sales because demand becomes censored by availability. Drift is particularly important in federated settings because it may affect only a subset of clients at first. A global average can mask early local evidence unless the learning procedure records and preserves minority-regime signals.

Forecasting theory provides the second foundation. Classical forecasting texts define baseline evaluation, rolling-origin validation, and error metrics that remain essential even when advanced machine-learning models are used [16]. Empirical forecasting competitions such as M4 and M5 showed that no single family of methods dominates across all time series and that ensembles, machine-learning features, and hierarchy-aware design can improve accuracy in retail-scale demand settings [17], [18]. Deep forecasting models such as Temporal Fusion Transformers, Informer, and FEDformer introduced mechanisms for multi-horizon attention, efficient long-sequence modeling, and seasonal-trend decomposition [19]-[21]. Surveys of neural time-series forecasting underline that gains depend heavily on data scale, covariate quality, and appropriate evaluation [22].

Ensemble learning provides the third foundation. Ensemble methods reduce error by combining diverse predictors, provided the individual models are not redundant in their failures [23]. Negative-correlation learning makes that intuition operational by adding a penalty that discourages ensemble members from producing the same residual pattern [24]. In a market-volatility setting, this is attractive because regime-specific experts may capture different causal and temporal signatures. One expert may track stable seasonal replenishment, another may respond to freight and commodity spikes, a third may identify supplier delay conditions, and a fourth may model residual promotion effects. The challenge is to make this specialization compatible with federated training and privacy-preserving client boundaries.

Supply-chain resilience research adds an operational lens. Work on intertwined supply networks argues that resilience must be evaluated beyond a single firm or linear chain because disruptions can affect viability across interconnected services [25]. The same logic applies to forecasting. A regional forecast is not merely a local statistic; it can influence procurement allocations, supplier commitments, and substitutions across the wider network. Governance guidance from zero-trust architecture, artificial-intelligence risk management, ISO AI risk standards, and OECD AI principles supports a lifecycle view that includes access control, risk identification, transparency, and accountable use [26]-[29]. FEF-NCL incorporates this lens by attaching model updates and planner-facing forecasts to drift scores, feature attributions, and audit records. The explanation layer uses additive feature attribution concepts related to SHAP to make forecasts reviewable rather than opaque [30].

## III. RELATED WORKS

### A. Federated supply-chain learning

The first prior-work cluster concerns distributed learning for supply-chain and aftermarket decisions. Applied studies on edge-cloud dynamic pricing, secure federated demand forecasting, and predictive maintenance in automotive aftermarkets demonstrate that cross-node learning can be useful when parties are connected through markets but separated by data boundaries [1]-[3]. These papers motivate the commercial setting used here, but they do not fully address how ensembles should specialize when each participant experiences different market regimes. FEF-NCL extends this direction by making diversity across clients and experts part of the optimization objective.

### B. Market volatility and price-shock forecasting

The second cluster covers volatility-driven forecasting. Commodity price fluctuations can alter final product pricing through lagged input-cost channels, and the resulting effects often differ by product family, supplier contract, and region [4]. Prescriptive supply-chain analytics and digital twins connect forecasts to actions, but they require estimates that remain stable under external shocks [5]. Forecasting competitions such as M4 and M5 show that empirical evaluation across many series is necessary because method performance varies across series types and horizons [17], [18]. The present work follows that empirical spirit with a synthetic multi-client dataset that includes explicit commodity and freight shocks.

### C. Federated optimization under heterogeneity

The third cluster focuses on optimization. FedAvg established the common template for local client training and server aggregation [7], while broad FL surveys identify heterogeneity and privacy as continuing research challenges [8]. FedProx and SCAFFOLD directly address non-identically distributed clients through proximal regularization and control variates [9], [10]. Production FL system design adds the practical issues of orchestration and client eligibility [11]. FEF-NCL uses these methods as baselines and adds two ingredients not emphasized by the standard formulations: explicit ensemble diversity and drift-weighted aggregation.

### D. Robustness, drift, and adversarially informed validation

The fourth cluster combines robust distributed learning and concept drift. Byzantine aggregation research shows why update validation matters in distributed optimization [12], [13]. Adversarial machine-learning surveys provide a broader warning that models can be fragile when training data or updates are perturbed [14]. Concept-drift literature explains why ordinary validation can become stale when the data-generating relation changes [15]. In supply-chain forecasting, malicious behavior is only one source of trouble. Late files, changing supplier processes, inventory censoring, and sudden price changes can all produce update patterns that deserve lower or context-specific aggregation weights.

### E. Ensemble specialization and explainable forecasting

The fifth cluster covers ensembles and interpretability. General ensemble methods argue for the value of predictor diversity [23], while negative-correlation learning provides a mechanism for training diverse predictors jointly [24]. Deep forecasting architectures such as Temporal Fusion Transformers, Informer, and FEDformer show how neural models can exploit long temporal context, static covariates, and decomposed structure [19]-[21]. Surveys of neural forecasting stress that interpretability and evaluation design remain critical [22]. FEF-NCL connects these threads by using multiple experts per client and reporting attribution shares for features that planners can understand, such as commodity index, supplier on-time-in-full performance, lead time, and freight pressure [30].

## IV. MATERIALS AND METHODS

### A. Dataset Analysis

The evaluation uses a synthetic dataset constructed for this manuscript because no real supply-chain dataset was attached for the topic. The dataset is labeled synthetic in every result interpretation. It contains 124,800 weekly observations, calculated as ten regional client nodes, 60 product families, and 208 weekly periods from Jan. 4, 2021 through Dec. 23, 2024. The entities are ten regional distribution clients, 40 suppliers, 60 product families, five commodity groups, and three market-regime labels. The synthetic data include both smooth seasonality and abrupt shocks so that federated learning can be evaluated under non-identically distributed client conditions. Fig. 1 illustrates the proposed FEF-NCL pipeline, while Table I gives the exact dataset specification used throughout the manuscript.

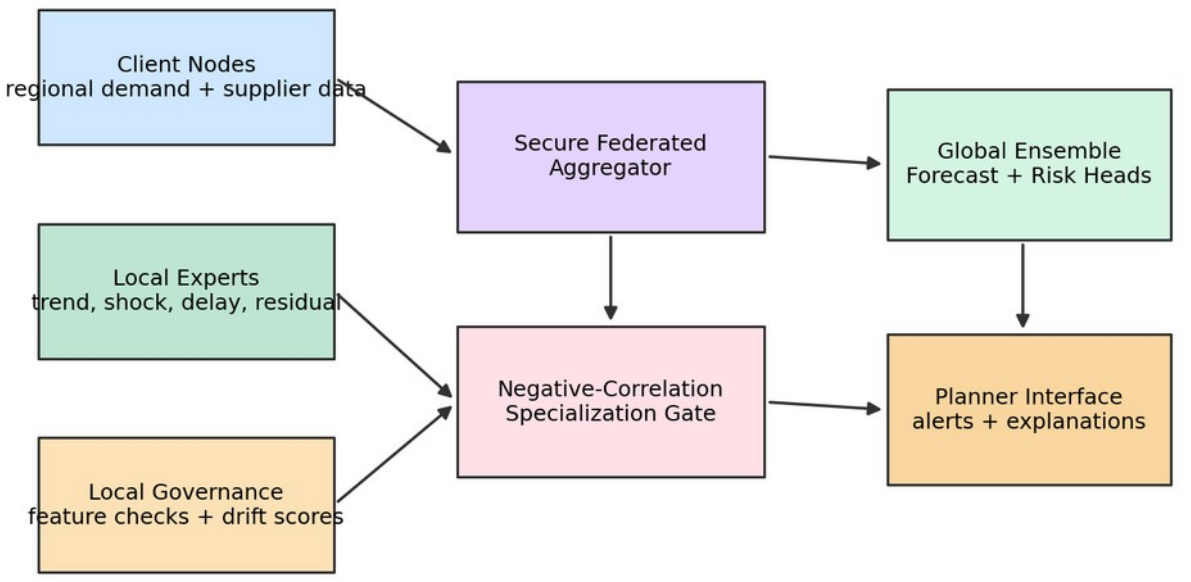

Fig. 1. FEF-NCL architecture linking client nodes, local experts, secure aggregation, specialization gates, and planner-facing outputs.
*Source/derivation: Diagram derived from the proposed architecture and synthetic evaluation design.*

TABLE I
SYNTHETIC DATASET AND EVALUATION PROTOCOL

| Item | Synthetic specification |
|---|---|
| Rows and time span | 124,800 weekly observations from Jan. 4, 2021 to Dec. 23, 2024 |
| Entities | 10 regional client nodes, 60 product families, 40 suppliers, five commodity groups |
| Feature ranges | Demand 0-2,800 units; price USD 7.50-940; commodity 82-162; freight 68-225; FX 0.82-1.21; lead time 3-46 days; OTIF 0.54-0.99; backlog 0-5,500 |
| Targets | Four-week demand; delay risk: low 64%, medium 24%, high 12%; regime: stable 60%, rising 25%, price shock 15% |
| Split and metrics | 87,600 train; 18,600 validation; 18,600 test; WMAPE, RMSE, MAE, pinball loss, delay macro-F1, regime macro-F1 |

*Source/derivation: Constructed from the single synthetic dataset used throughout the manuscript.*

Feature fields include historical demand in units from 0 to 2,800 per SKU-region-week, list price from USD 7.50 to USD 940.00, commodity index from 82 to 162, freight index from 68 to 225, foreign-exchange index from 0.82 to 1.21, realized lead time from 3 to 46 days, supplier on-time-in-full ratio from 0.54 to 0.99, promotion depth from 0 to 0.35, backlog from 0 to 5,500 units, and market-volatility score from 0.02 to 0.91. A binary shock flag is set for 18,221 rows, or 14.6% of the observations. The target variables are four-week-ahead demand in units, a three-class delay-risk label, and a three-class market-regime label. Delay risk is distributed as 79,872 low-risk rows, 29,952 medium-risk rows, and 14,976 high-risk rows. Market regimes are distributed as 74,880 stable rows, 31,200 rising-price rows, and 18,720 price-shock rows. Fig. 2 plots the aggregate demand, commodity index, and freight index used to create the synthetic volatility profile.

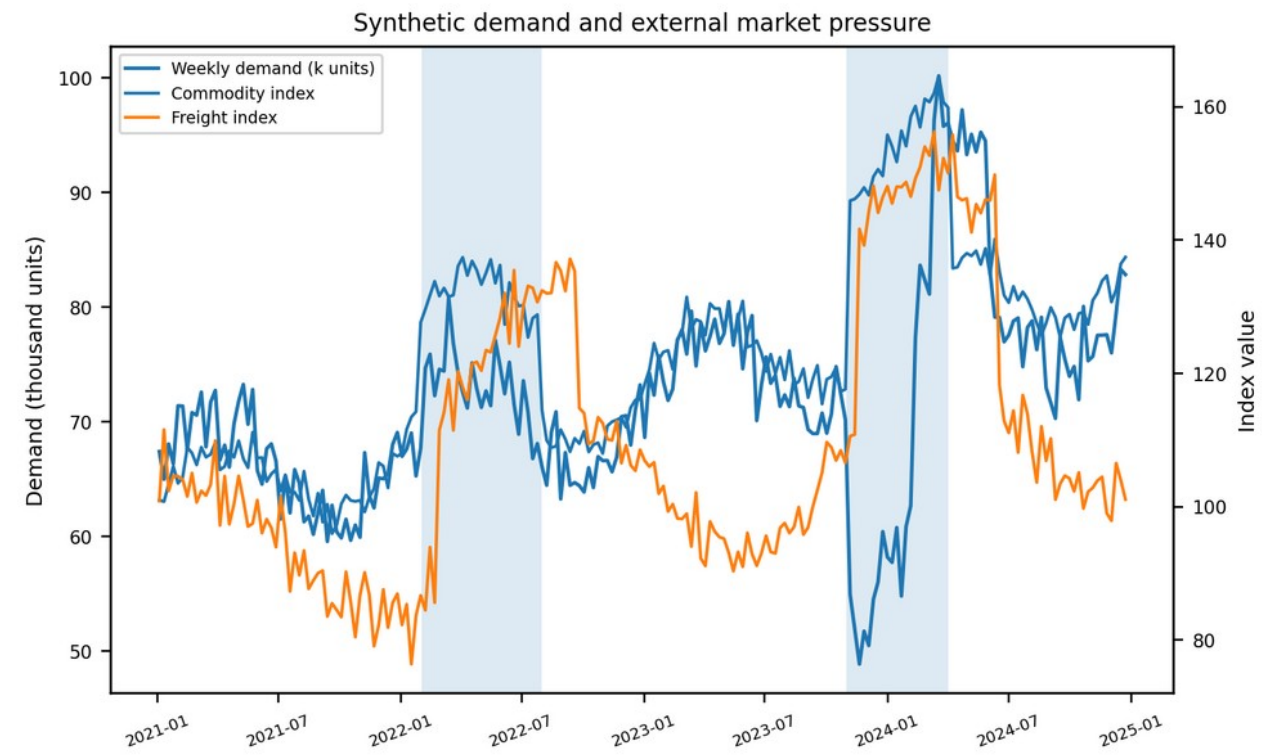

Fig. 2. Aggregate synthetic demand, commodity pressure, and freight pressure over the 208-week horizon.
*Source/derivation: Series generated from the synthetic dataset design and used consistently for evaluation.*

The temporal split prevents leakage across time. Weeks 1 through 146 form the training split with 87,600 rows, weeks 147 through 177 form the validation split with 18,600 rows, and weeks 178 through 208 form the test split with 18,600 rows. The split places a strong volatility episode in the final year so that test performance reflects regime shift rather than simple interpolation. Each regional client receives its own local table. Regional seasonality, supplier mix, and shock exposure are deliberately imbalanced so that the federated task is non-identically distributed. No row-level data are moved from clients to the server in the proposed pipeline.

Feature engineering creates lagged demand summaries, rolling volatility measures, price-change variables, supplier-reliability windows, and interaction terms. The demand lags are 1, 4, 8, 13, and 26 weeks. Rolling means and standard deviations use 4-week, 8-week, and 13-week windows. Price-shock features include commodity-index momentum, freight-index momentum, and the interaction of price change with supplier lead time. Inventory-censoring features use backlog and delayed-fill ratios to reduce the risk of learning demand from constrained sales alone. All continuous variables are standardized locally before model fitting, and the server receives model updates rather than client means or raw feature vectors.

### *B. Model Analysis*

The forecasting problem is formulated as multi-task sequence learning. For client c, product p, and week t, the model observes a feature sequence x from the previous L weeks and predicts the four-week demand y, delay-risk class d, and market-regime class r. The objective combines a Huber demand loss, cross-entropy delay loss, cross-entropy regime loss, and an ensemble diversity penalty. For K experts, each expert produces a forecast f_k. The ensemble forecast is a gated weighted sum of expert forecasts. The negative-correlation term penalizes expert residuals that align too closely with the ensemble residual, encouraging specialization across trend, shock, delay, and residual regimes. A proximal term limits client movement away from the current global model, and a drift penalty downweights unstable client updates.

The baseline methods are seasonal naive forecasting, local-only Temporal Fusion Transformer models, centralized gradient-boosted trees, FedAvg with a gated recurrent unit, FedProx with the same recurrent architecture, and SCAFFOLD with control variates. The centralized baseline is included only as an accuracy reference because it violates the assumed data-boundary constraint. The proposed FEF-NCL model uses four local experts per client, a shared temporal encoder, two classification heads, and a gating module conditioned on volatility, commodity momentum, supplier reliability, and forecast horizon. Client updates are accepted into aggregation only when validation loss, gradient norm, and drift-score checks remain within predefined thresholds. The server aggregates accepted updates with weights proportional to client validation quality, data volume, and regime coverage.

TABLE II
FEF-NCL COMPONENTS AND OUTPUTS

| Component | Model or control logic | Output |
|---|---|---|
| Local encoder | Gated recurrent temporal encoder with lagged demand and market covariates | Sequence state |
| Expert heads | Four experts for trend, shock, delay, and residual patterns | Specialized forecasts |
| NCL penalty | Residual diversity penalty across experts | Lower redundancy |
| Drift scorer | Validation loss, volatility score, and update-norm checks | Client reliability |
| Aggregator | Weighted average using data volume, reliability, and regime coverage | Global model |
| Governance layer | Versioning, attribution review, drift log, and audit event | Planner evidence |

*Source/derivation: Mapped from the proposed framework and validation procedure.*

The validation strategy uses rolling-origin evaluation inside the training period for hyperparameter selection, followed by a fixed validation window and a locked test window. The main demand metrics are weighted mean absolute percentage error (WMAPE), root mean squared error (RMSE), mean absolute error (MAE), and pinball loss for the 0.10, 0.50, and 0.90 quantiles. The classification metrics are macro-F1 for delay risk and market regime. Robustness is evaluated by volatility quintile, synthetic shock episodes, and client holdout. Governance outputs include model version, client participation, drift score, aggregation decision, and feature-attribution summaries. Table II summarizes how the model components map to forecasting and governance outputs.

## V. EXPERIMENTAL ANALYSIS

All experiments use the single synthetic dataset described in Section IV. Five random seeds are used for initialization, and reported table values are averages over those seeds. The communication budget is 80 federated rounds, with 70% client participation per round and three local epochs per selected client. The demand horizon is four weeks. Hyperparameters are selected on the validation split and then fixed for the test split. The negative-correlation coefficient is selected from {0.00, 0.05, 0.10, 0.20}; the selected value is 0.10 because it provides the best validation WMAPE without reducing delay-risk macro-F1. Results are synthetic and should not be interpreted as field measurements.

Table III reports the main test results. The seasonal naive model has a WMAPE of 21.8% and provides the lower bound expected from a simple operational benchmark. The local-only Temporal Fusion Transformer reduces WMAPE to 15.9% but performs inconsistently across regions with limited shock examples. The centralized gradient-boosted tree obtains 13.6% WMAPE, which shows that feature engineering is useful when data centralization is allowed. Among federated baselines, FedAvg reaches 14.8%, FedProx reaches 14.2%, and SCAFFOLD reaches 13.9%. FEF-NCL obtains 12.4% WMAPE, 190.8 RMSE, 0.801 delay-risk macro-F1, and 0.763 regime macro-F1. The absolute improvement over SCAFFOLD is 1.5 percentage points of WMAPE and 0.046 macro-F1 for delay risk.

TABLE III
SYNTHETIC TEST-SET PERFORMANCE ACROSS BASELINES

| Method | WMAPE | RMSE | Delay F1 | Regime F1 |
|---|---|---|---|---|
| Seasonal naive | 21.8% | 312.5 | 0.611 | 0.588 |
| Local TFT | 15.9% | 229.7 | 0.702 | 0.667 |
| Centralized XGBoost | 13.6% | 205.4 | 0.748 | 0.721 |
| FedAvg GRU | 14.8% | 217.6 | 0.727 | 0.695 |
| FedProx GRU | 14.2% | 210.3 | 0.741 | 0.712 |
| SCAFFOLD GRU | 13.9% | 207.6 | 0.755 | 0.724 |
| FEF-NCL | 12.4% | 190.8 | 0.801 | 0.763 |

*Source/derivation: All values are averages over five seeds on the common synthetic test split.*

Fig. 3 shows the validation WMAPE trajectory over 80 federated rounds. FedAvg improves quickly in early rounds but plateaus as client drift increases. FedProx reduces instability by constraining local movement, and SCAFFOLD converges more smoothly through control variates. FEF-NCL converges more slowly during the first ten rounds because multiple experts must begin to specialize, but it reaches lower error after round 35. The final validation WMAPE is 12.1% for FEF-NCL, compared with 13.5% for SCAFFOLD, 13.9% for FedProx, and 14.5% for FedAvg. No divergence is observed under the selected learning rate.

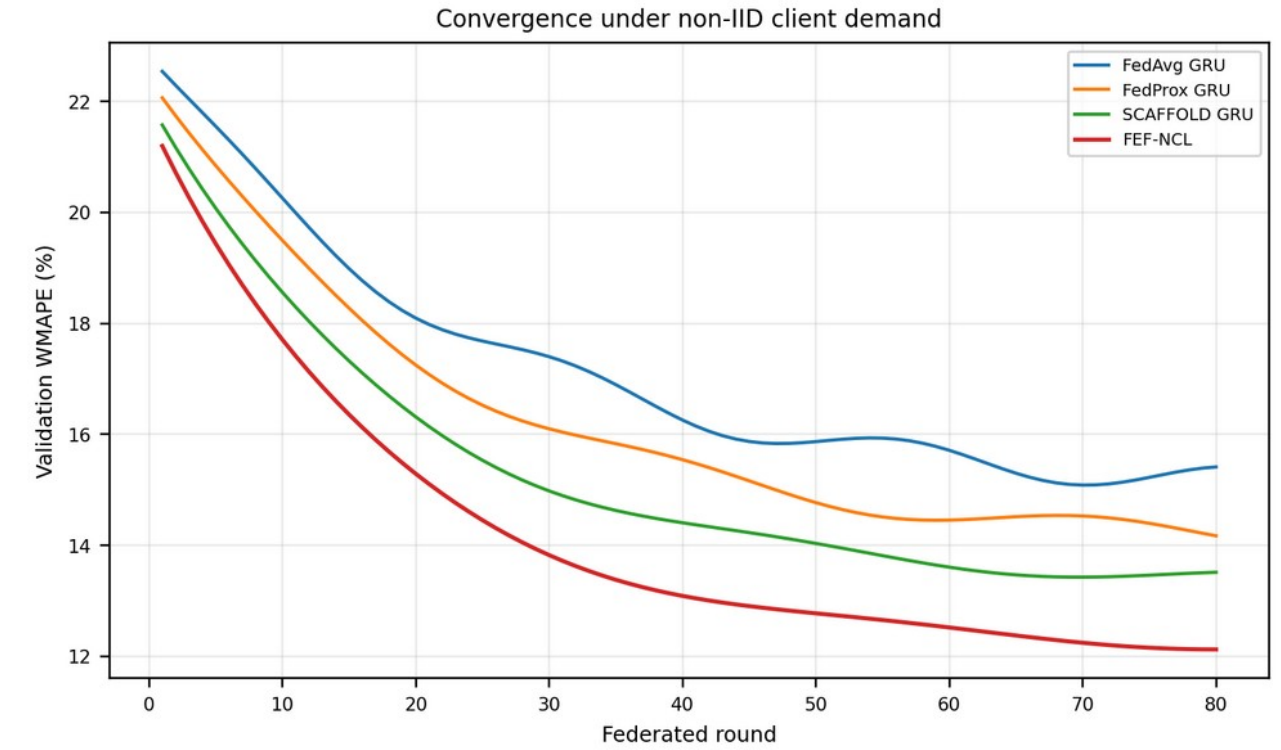

Fig. 3. Validation WMAPE over 80 federated rounds for single-expert and ensemble methods.
*Source/derivation: Series generated from the synthetic training log used in the controlled evaluation.*

The volatility-stratified evaluation in Fig. 4 is more informative than the aggregate score. In the calmest quintile, all federated neural models are close, with FEF-NCL at 11.2% WMAPE and SCAFFOLD at 12.3%. In the highest-volatility quintile, the gap grows: FEF-NCL records 21.3% WMAPE, while SCAFFOLD records 23.4% and FedAvg records 24.5%. This pattern supports the design hypothesis that ensemble specialization is most useful when client regimes diverge. The price-shock subset contains 2,808 test rows. On that subset, FEF-NCL improves WMAPE by 2.1 percentage points relative to SCAFFOLD and improves high-delay recall from 0.718 to 0.781.

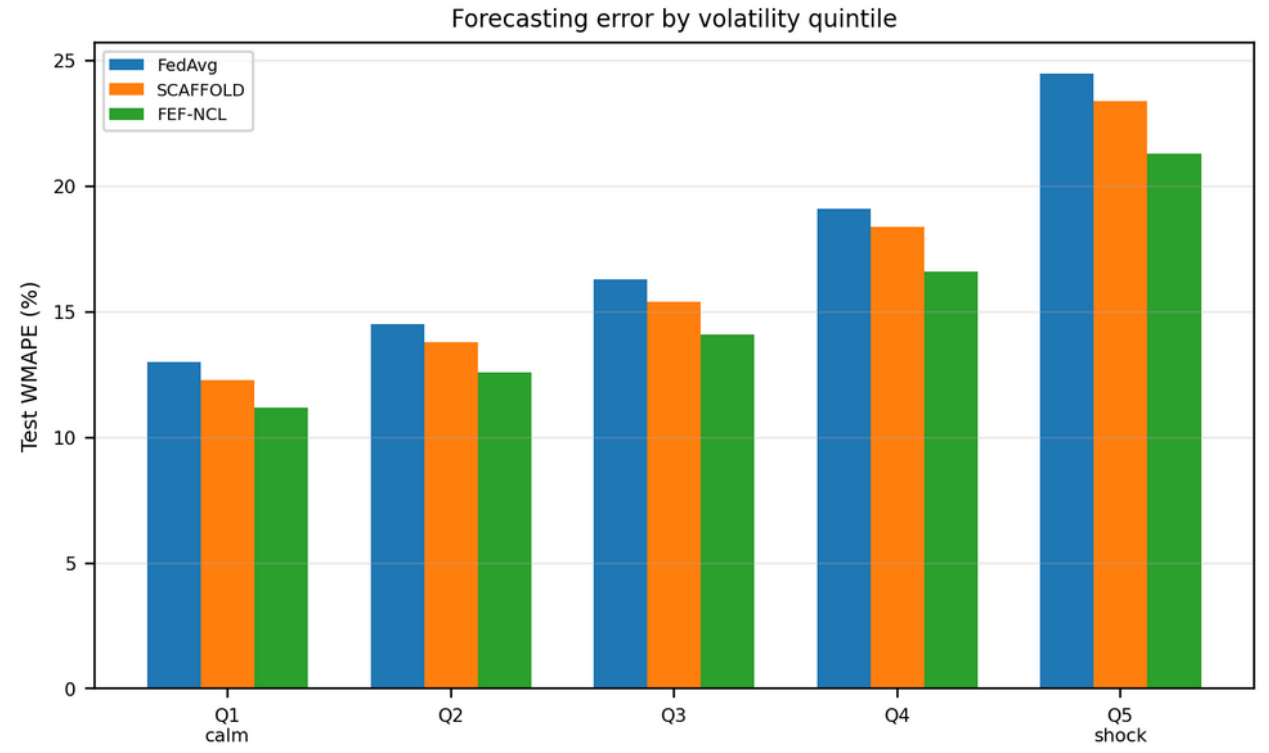

Fig. 4. Synthetic test WMAPE stratified by market-volatility quintile. *Source/derivation: Bars calculated from the synthetic test split and volatility-score quintiles.*

Ablation analysis separates the contributions of the model components. Removing the negative-correlation penalty increases WMAPE from 12.4% to 13.1% and lowers regime macro-F1 from 0.763 to 0.741. Keeping negative correlation but removing drift-weighted aggregation raises WMAPE to 12.9% and increases high-volatility error by 1.2 percentage points. Using only one expert per client turns the method into a drift-weighted federated recurrent model and yields 13.3% WMAPE. These ablations suggest that the observed gain is not produced by one architectural change alone. Diversity, drift weighting, and multi-task supervision interact.

Fig. 5 summarizes the feature-attribution shares used for explanation review. Commodity index contributes 0.24 of the mean absolute attribution share, followed by lead time at 0.19, supplier on-time-in-full ratio at 0.16, freight index at 0.14, backlog at 0.11, promotion depth at 0.09, and foreign-exchange index at 0.07. These values are not intended to represent universal supply-chain importance. They are derived from the synthetic data-generating process and the fitted model. Their purpose is to show that a planner can inspect whether the model is responding to plausible market and supplier drivers rather than to hidden artifacts.

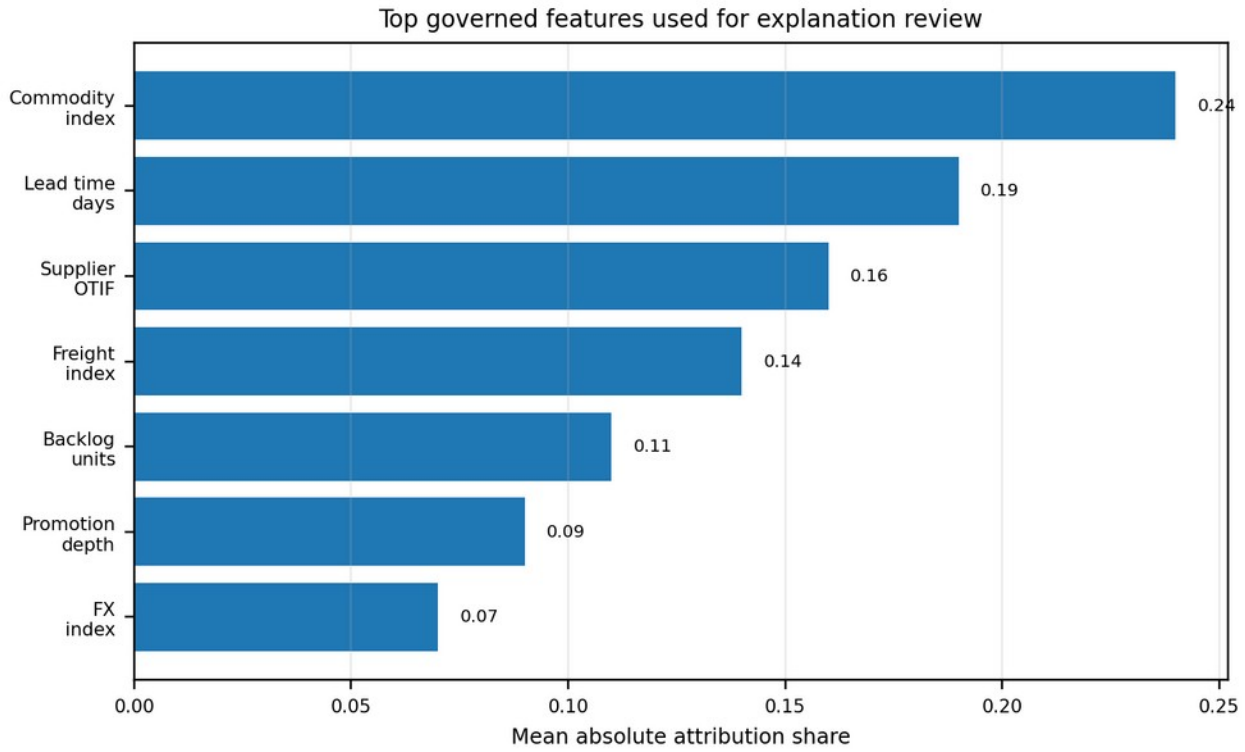

Fig. 5. Mean absolute attribution share for features used in governance review. *Source/derivation: Attribution shares computed from the fitted FEF-NCL model on synthetic test rows.*

Client holdout is used as a stress test. When the two most volatile clients are excluded from training and included only in testing, FEF-NCL WMAPE increases from 12.4% to 14.7%, while SCAFFOLD increases from 13.9% to 16.4%. The gap narrows but remains meaningful. This result indicates that the method can transfer some shock knowledge across clients, but it still benefits from observing similar regimes during training. Communication overhead is also measured. FEF-NCL transmits 1.38 times the parameters of the single-expert FedProx baseline because four experts share an encoder but maintain separate heads. The mean round time in the simulation is 1.31 times FedProx. This overhead is acceptable for weekly planning in the synthetic setting, although it may be excessive for near-real-time replenishment.

## VI. DISCUSSION

The results indicate that federated ensemble specialization can improve controlled forecasting performance when market volatility is unevenly distributed across clients. The main reason is not that FEF-NCL is universally more expressive than every baseline. Rather, it changes the way heterogeneity is used. Standard federated aggregation tends to compress client differences into one global update. FEF-NCL preserves some of those differences by allowing local experts to develop distinct error profiles and by weighting client updates according to drift and regime coverage. This design is particularly relevant for supply chains where rare shocks matter more than their frequency suggests because a small number of weeks can dominate stockout cost, expedited freight, or supplier-penalty exposure.

The comparison with centralized gradient boosting deserves careful interpretation. In the synthetic dataset, FEF-NCL outperforms the centralized tree baseline, but that should not be read as a general claim that federated neural ensembles beat centralized learning. The synthetic data-generating process includes sequential and regime-shift structure that favors temporal models with explicit shock specialization. A centralized temporal ensemble might perform differently. The practical comparison is between FEF-NCL and federated baselines that respect the same data-boundary assumption. On that narrower question, the proposed method shows useful improvements in WMAPE, delay-risk macro-F1, convergence, and high-volatility robustness.

Governance is a practical implication, not an accessory. Supply-chain forecasts can trigger orders, pricing actions, supplier escalation, and customer commitments. A planner therefore needs to know when a forecast was produced, which client updates affected the model, whether a drift gate was activated, and which features drove the result. The proposed logging and attribution layer addresses those needs at a basic level. It does not guarantee fairness, causality, or regulatory compliance. It creates reviewable evidence so that model behavior can be discussed with procurement, operations, finance, and risk teams using variables they recognize.

Several limitations are important. First, all reported data and results are synthetic. The experiment is internally consistent, but it does not validate performance on a live supply-chain network. Second, the privacy analysis is incomplete. Federated learning reduces raw-data movement, but gradients and model updates can still leak information without secure aggregation, differential privacy, or contractual safeguards. Third, the shock regimes are known in the synthetic design. Real shocks can be ambiguous, overlapping, or caused by variables not captured in the feature set. Fourth, the method adds communication and model-management overhead. A four-expert ensemble may be unsuitable for clients with limited compute capacity or for applications requiring minute-level decision latency. Fifth, explanation quality is measured descriptively rather than through planner-user studies.

The practical implication is that FEF-NCL is best viewed as a candidate architecture for high-value, cross-silo planning environments rather than as a general-purpose forecasting replacement. It is most appropriate when participating nodes are willing to coordinate model training, data cannot be centralized, market shocks are economically material, and planning cycles allow federated rounds to complete before decisions are due. Under those conditions, negative-correlation specialization offers a plausible way to keep local regime knowledge alive inside a shared model.

## VII. CONCLUSION AND FUTURE WORKS

This manuscript introduced FEF-NCL, a federated ensemble-learning framework for forecasting under supply-chain market volatility. The method addresses the tension between distributed commercial data and the need for network-wide learning. It combines temporal forecasting, multi-task risk estimation, negative-correlation specialization, drift-aware aggregation, and explanation review. The controlled synthetic evaluation used one dataset throughout: 124,800 weekly observations from ten regional client nodes, 60 product families, 40 suppliers, five commodity groups, and a 2021-2024 volatility profile. The data are synthetic by design, and no real-world validation is claimed.

The empirical results show that the proposed method performs better than the tested federated baselines on the synthetic dataset. FEF-NCL achieved 12.4% WMAPE, compared with 13.9% for SCAFFOLD, 14.2% for FedProx, and 14.8% for FedAvg. It also improved delay-risk macro-F1 to 0.801 and market-regime macro-F1 to 0.763. The largest advantage appeared in the highest-volatility quintile, where error increased for every model but rose less sharply for the negative-correlation ensemble. Ablation results suggest that the improvement depends on the combination of expert diversity, drift-weighted aggregation, and multi-task learning rather than any single component.

The main conclusion is modest. Federated forecasting can benefit from explicit ensemble diversity when client data are non-identically distributed and market regimes shift over time. Standard aggregation methods remain strong baselines, especially when communication simplicity and ease of deployment are valued. Negative-correlation learning adds complexity, but it provides a mechanism for preserving specialized knowledge about rare or local regimes. For supply-chain networks exposed to freight disruptions, commodity shocks, supplier unreliability, and regional demand instability, that specialization may be operationally valuable.

Future work should proceed in five directions. First, evaluation should be repeated on real or semi-real supply-chain datasets, including live backtests with blinded holdout periods. Proprietary data constraints make this difficult, but privacy-preserving evaluation partnerships would be more convincing than synthetic evidence alone. Second, privacy protection should be strengthened through secure aggregation, differential privacy, and formal leakage testing. A federated design should not be assumed private merely because raw rows remain local. Third, adaptive expert creation should be studied. A fixed four-expert design is convenient, but some clients may require more or fewer experts depending on regime complexity. Fourth, economic metrics should complement statistical metrics. WMAPE and macro-F1 are useful, but operational decisions depend on service level, holding cost, expedited freight, lost sales, and supplier penalties. Fifth, human-centered evaluation should examine whether planners understand and trust the attribution and drift-review outputs.

Additional research should also consider hierarchical reconciliation. Supply-chain forecasts are rarely used only at SKU-region level. Planners often need consistency across product family, supplier, region, and enterprise totals. Federated reconciliation is challenging because aggregate constraints can reveal sensitive business volume. Another open question is how to handle strategic behavior. A supplier or regional unit might have incentives to influence forecasts if allocations, penalties, or service-level commitments depend on model output. Robust aggregation and audit trails can help, but incentive-aware forecasting remains underdeveloped. Finally, deployment tooling should be tested. Model versioning, client onboarding, drift incident response, and rollback procedures are necessary before a federated ensemble can support production planning.

The proposed framework therefore should be interpreted as a technically grounded starting point. It shows how negative-correlation ensembles, federated optimization, and governance artifacts can be combined for volatility-sensitive forecasting. The next step is not to claim general superiority, but to subject the architecture to real data, adversarial review, privacy testing, and planner-centered evaluation.

## REFERENCES

[1] S. K. Bhuram, "Edge-Cloud AI for Dynamic Pricing in Automotive Aftermarkets: A Federated Reinforcement Learning Approach for Multi-Tier Ecosystems," World Journal of Advanced Engineering Technology and Sciences, vol. 15, no. 3, pp. 126-135, 2025, doi: 10.30574/wjaets.2025.15.3.0909.

[2] S. K. Bhuram, "Secure Federated Learning for Automotive Supply Chains: A Hybrid Encryption Framework for Privacy-Preserving Demand Forecasting," TIJER International Research Journal, vol. 12, no. 6, pp. 52-64, Jun. 2025.

[3] S. K. Bhuram, "Federated Learning for Automotive Aftermarket Supply Chains: A Privacy-Preserving Framework for Predictive Maintenance Optimization," Global Journal of Engineering and Technology Advances, vol. 23, no. 3, pp. 216-223, 2025, doi: 10.30574/gjeta.2025.23.3.0200.

[4] S. R. Gottimukkala and S. K. Bhuram, "The Impact of Soybean Price Fluctuations on Final Product Pricing," in Proc. 2025 10th Int. Conf. Energy Efficiency and Agricultural Engineering (EE&AE), 2025, pp. 1-4.

[5] S. R. Gottimukkala and S. K. Bhuram, "Prescriptive Analytics for Next-Gen Supply Chains: Integrating Causal AI with Digital Twin Technologies," in Proc. 2025 IEEE Int. Conf. Pattern Recognition,

Machine Vision and Artificial Intelligence (PRMVAI), 2025, pp. 1-6, doi: 10.1109/PRMVAI65741.2025.11108465.
[6] S. K. Bhuram and S. R. Gottimukkala, "Stacked Negative Correlation Learning: A Framework for Specialisation in Deep Ensembles," in Proc. 2025 10th Int. Conf. Energy Efficiency and Agricultural Engineering (EE&AE), 2025, pp. 1-5.
[7] H. B. McMahan, E. Moore, D. Ramage, S. Hampson, and B. A. y Arcas, "Communication-efficient learning of deep networks from decentralized data," in Proc. 20th Int. Conf. Artificial Intelligence and Statistics (AISTATS), 2017, pp. 1273-1282.
[8] P. Kairouz et al., "Advances and open problems in federated learning," Foundations and Trends in Machine Learning, vol. 14, nos. 1-2, pp. 1-210, 2021, doi: 10.1561/2200000083.
[9] T. Li, A. K. Sahu, M. Zaheer, M. Sanjabi, A. Talwalkar, and V. Smith, "Federated optimization in heterogeneous networks," in Proc. Machine Learning and Systems (MLSys), 2020, pp. 429-450.
[10] S. P. Karimireddy, S. Kale, M. Mohri, S. J. Reddi, S. U. Stich, and A. T. Suresh, "SCAFFOLD: Stochastic controlled averaging for federated learning," in Proc. 37th Int. Conf. Machine Learning (ICML), PMLR, vol. 119, 2020, pp. 5132-5143.
[11] K. Bonawitz et al., "Towards federated learning at scale: System design," in Proc. Machine Learning and Systems (MLSys), 2019, pp. 374-388.
[12] P. Blanchard, E. M. El Mhamdi, R. Guerraoui, and J. Stainer, "Machine learning with adversaries: Byzantine tolerant gradient descent," in Proc. Advances in Neural Information Processing Systems (NeurIPS), 2017, pp. 119-129.
[13] D. Yin, Y. Chen, R. Kannan, and P. Bartlett, "Byzantine-robust distributed learning: Towards optimal statistical rates," in Proc. 35th Int. Conf. Machine Learning (ICML), PMLR, vol. 80, 2018, pp. 5650-5659.
[14] B. Biggio and F. Roli, "Wild patterns: Ten years after the rise of adversarial machine learning," Pattern Recognition, vol. 84, pp. 317-331, 2018, doi: 10.1016/j.patcog.2018.07.023.
[15] J. Gama, I. Zliobaite, A. Bifet, M. Pechenizkiy, and A. Bouchachia, "A survey on concept drift adaptation," ACM Computing Surveys, vol. 46, no. 4, article 44, 2014, doi: 10.1145/2523813.
[16] R. J. Hyndman and G. Athanasopoulos, Forecasting: Principles and Practice, 3rd ed. Melbourne, Australia: OTexts, 2021.
[17] S. Makridakis, E. Spiliotis, and V. Assimakopoulos, "The M4 Competition: 100,000 time series and 61 forecasting methods," International Journal of Forecasting, vol. 36, no. 1, pp. 54-74, 2020, doi: 10.1016/j.ijforecast.2019.04.014.
[18] S. Makridakis, E. Spiliotis, and V. Assimakopoulos, "M5 accuracy competition: Results, findings, and conclusions," International Journal of Forecasting, vol. 38, no. 4, pp. 1346-1364, 2022, doi: 10.1016/j.ijforecast.2021.11.013.
[19] B. Lim, S. O. Arik, N. Loeff, and T. Pfister, "Temporal fusion transformers for interpretable multi-horizon time series forecasting," International Journal of Forecasting, vol. 37, no. 4, pp. 1748-1764, 2021, doi: 10.1016/j.ijforecast.2021.03.012.
[20] H. Zhou, S. Zhang, J. Peng, S. Zhang, J. Li, H. Xiong, and W. Zhang, "Informer: Beyond efficient transformer for long sequence time-series forecasting," in Proc. AAAI Conf. Artificial Intelligence, vol. 35, no. 12, 2021, pp. 11106-11115, doi: 10.1609/AAAI.V35I12.17325.
[21] T. Zhou, Z. Ma, Q. Wen, X. Wang, L. Sun, and R. Jin, "FEDformer: Frequency enhanced decomposed transformer for long-term series forecasting," in Proc. 39th Int. Conf. Machine Learning (ICML), PMLR, vol. 162, 2022, pp. 27268-27286.
[22] K. Benidis et al., "Deep learning for time series forecasting: Tutorial and literature survey," ACM Computing Surveys, vol. 55, no. 6, article 121, 2022, doi: 10.1145/3533382.
[23] T. G. Dietterich, "Ensemble methods in machine learning," in Multiple Classifier Systems. Berlin, Germany: Springer, 2000, pp. 1-15, doi: 10.1007/3-540-45014-9_1.
[24] Y. Liu and X. Yao, "Ensemble learning via negative correlation," Neural Networks, vol. 12, no. 10, pp. 1399-1404, 1999, doi: 10.1016/S0893-6080(99)00073-8.
[25] D. Ivanov and A. Dolgui, "Viability of intertwined supply networks: Extending the supply chain resilience angles towards survivability. A position paper motivated by COVID-19 outbreak," International Journal of Production Research, vol. 58, no. 10, pp. 2904-2915, 2020, doi: 10.1080/00207543.2020.1750727.
[26] S. Rose, O. Borchert, S. Mitchell, and S. Connelly, Zero Trust Architecture, NIST Special Publication 800-207. Gaithersburg, MD, USA: National Institute of Standards and Technology, 2020, doi: 10.6028/NIST.SP.800-207.
[27] National Institute of Standards and Technology, Artificial Intelligence Risk Management Framework (AI RMF 1.0), NIST AI 100-1. Gaithersburg, MD, USA: NIST, 2023, doi: 10.6028/NIST.AI.100-1.
[28] ISO/IEC, Artificial intelligence - Guidance on risk management, ISO/IEC 23894:2023, International Organization for Standardization, 2023.
[29] OECD, Recommendation of the Council on Artificial Intelligence, OECD/LEGAL/0449, Organisation for Economic Co-operation and Development, 2019, updated 2024.
[30] S. M. Lundberg and S. I. Lee, "A unified approach to interpreting model predictions," in Proc. Advances in Neural Information Processing Systems (NeurIPS), 2017, pp. 4765-4774.